\documentclass{article}
\usepackage{arxiv}
\usepackage{natbib}
\usepackage[utf8]{inputenc} % allow utf-8 input
\usepackage[T1]{fontenc}    % use 8-bit T1 fonts
\usepackage{hyperref}       % hyperlinks
\usepackage{url}            % simple URL typesetting
\usepackage{booktabs}       % professional-quality tables
\usepackage{amsfonts}       % blackboard math symbols
\usepackage{nicefrac}       % compact symbols for 1/2, etc.
\usepackage{microtype}      % microtypography
\usepackage{lipsum}		% Can be removed after putting your text content
\usepackage{graphicx}
\usepackage{doi}
\usepackage{color}
\usepackage{researchpack}
\usepackage[capitalise,nameinlink]{cleveref}
\usepackage{booktabs}
\usepackage{enumitem}
\usepackage{xspace}
\usepackage{multirow}
\usepackage{floatrow}
\usepackage[T1]{fontenc}
\usepackage{graphicx}
\usepackage[T1]{fontenc}
\usepackage{hyperref}
\usepackage{booktabs}
\usepackage{wrapfig}
\usepackage{amsmath, amssymb}
\usepackage[capitalise,nameinlink]{cleveref}

\usepackage{graphicx}
\usepackage{xcolor}
\usepackage[most]{tcolorbox}
\usepackage[misc]{ifsym}
\usepackage{multicol, multirow}
\usepackage{listings}
\usepackage{tikz}

\hypersetup{
    colorlinks=true,
    linkcolor=blue,
    citecolor=blue,
    urlcolor=blue,
}

\usepackage{xspace}
\newcommand{\MEdGellan}{{\sc MedGellan}\xspace}
\newcommand{\assistant}{{\sc assistant LLM}\xspace}
\newcommand{\physician}{{\sc physician}\xspace}
\newcommand{\physicianLM}{{\sc physician LLM}\xspace}
\newfloatcommand{capbtabbox}{table}[][\FBwidth]

\hypersetup{
    colorlinks=true,
    citecolor=blue,
    linkcolor=blue,
    filecolor=blue,
    urlcolor=blue,
}
\date{}

\title{\MEdGellan: LLM-Generated Medical Guidance to Support Physicians in Diagnosis}

\author{Debodeep Banerjee\\
	DI, University of Pisa\\
        DISI, University of Trento\\
	\And
        Burcu Sayin\\
    	DI, University of Pisa\\
        DISI, University of Trento\\
    \And
	   Stefano Teso  \\
	   CIMeC, university of Trento\\
       DISI, University of Trento\\
	\AND
	  Andrea Passerini \\
	  DISI, University of Trento \\
}

\renewcommand{\shorttitle}{\MEdGellan}

\begin{document}
\maketitle
\begin{abstract}
    Medical decision-making is a critical task, where errors can result in serious, potentially life-threatening consequences. While full automation remains challenging, hybrid frameworks that combine machine intelligence with human oversight offer a practical alternative. In this paper, we present \MEdGellan, a lightweight, annotation-free framework that uses a Large Language Model (LLM) to generate clinical guidance from raw medical records, which is then used by a physician to predict diagnoses. \MEdGellan uses a Bayesian-inspired prompting strategy that respects the temporal order of clinical data. Preliminary experiments show that the guidance generated by the LLM with \MEdGellan improves diagnostic performance, particularly in recall and $F_1$ score.
\end{abstract}

\newtcolorbox[auto counter, number within=section]{promptbox}[2][]{%
  colback=yellow!10,
  colframe=cyan!80!black,
  coltitle=black,
  fonttitle=\bfseries,
  title=Prompt for generating \textit{Guidance}~#2, #1,
  width=0.95\textwidth,
  label=box:#2,
}


\section{Introduction}
\label{sec:intro}

Medical diagnosis is a critical component of a patient's care, and accurately determining the diagnosis at the time of discharge from the hospital is one of a physician’s key responsibilities. Researchers explored automating this process by predicting clinical codes associated with a patient’s diagnosis \citep{de1998hierarchical,edin2023automated,baksi-etal-2025-medcoder,boyle2023automated, edin2023automated}. However, given the high-stakes nature of such decisions, it is not advisable to rely solely on machine-generated outputs ~\citep{Canada2019, AIact2021}. 

% \AP{I would rather write something like this "A popular strategy to address the risk of purely machine-generated
% predictions is {\em learning to defer}, in which the machine decides
% whether to output a prediction on a given instance or defer it to a
% human expert. (refs)" because also in our case the final decision is left to the human}

A popular strategy to mitigate the risk of purely machine generated predictions are {\em learning to defer} \citep{madras2018predict, mozannar2020consistent, keswani2022designing, verma2022calibrated, liu2022incorporating} and {\em learning to compliment} \citep{wilder2021learning}. However, in this case, when the machine predicts, the human either remains unassisted when they make the decision, or remains totally unaware of the system when the machine takes the decision.
Previous work explored hybrid human-machine medical decision-making scenarios \citet{banerjee2024learning} recognizes this phenomenon as \textit{separation of responsibilities} and argues that it is suboptimal, as one of the two agents always remains un-assisted while making a decision.
Evidently, this issue can compromise the reliability and efficiency of any decision-making system.

The problem discussed above results in a vacuum on the joint contribution of human and machine for a reliable decision-making framework. One plausible solution is to utilize machine intelligence as a helping hand to the human and to keep the human as the final decision maker. \citet{banerjee2024learning} offered a similar solution by \textit{finetuning} a vision language model (VLM) for generating radiology reports.
Nevertheless, finetuning an LLM/ VLM may turn out to be computationally costly.

\begin{figure*}[!t]
    \centering
    \begin{tabular}{ccc}
        \includegraphics[height=10em]{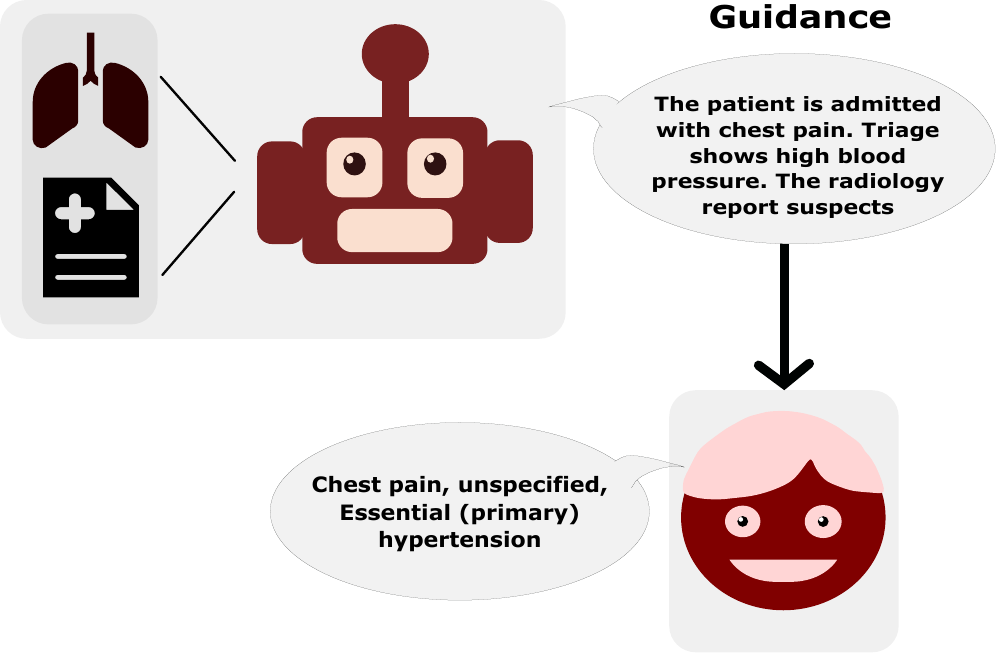}
        & 
        \begin{tikzpicture}
            \draw[dotted, ultra thick] (0,0) -- (0,10em);
        \end{tikzpicture}
        &
        \includegraphics[height=10em]{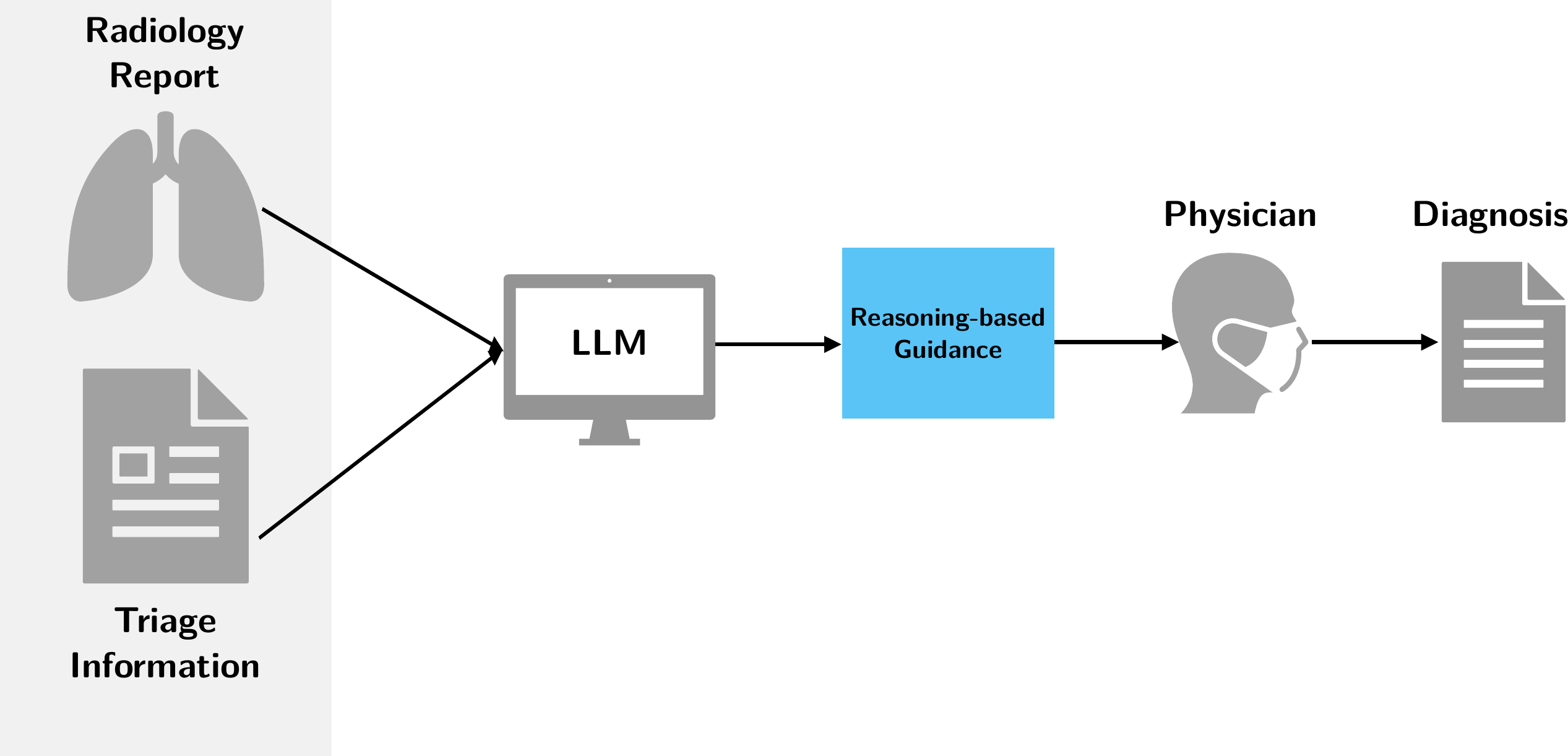}
    \end{tabular}
    \caption{\textbf{Left}: Simple illustration of \MEdGellan usage.
    \textbf{Right:} An end-to-end illustration of how \textit{guidance} generated by \MEdGellan is used to predict diagnosis.}
    \label{fig:pipeline}
\end{figure*}
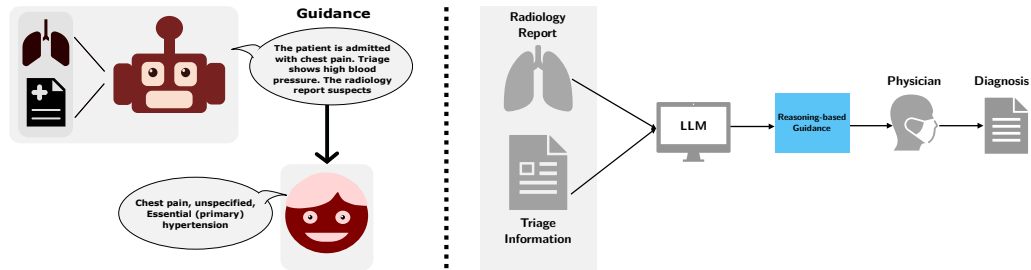

Based on these insights, we propose \MEdGellan, a novel pipeline where an \assistant is employed to analyze and provide \textit{guidance} for diagnosing a patient's health condition and in the later stage, a doctor, instead of studying the raw data or the electronic health record (EHR), can take advantage of the \textit{guidance} in order to make the final diagnosis (see \Cref{fig:pipeline}-left).

We demonstrate that with appropriate prompting strategies, LLMs are capable of generating high-quality guidance that can be equally helpful for the physician to make nuanced predictions of medical diagnosis. \MEdGellan requires no finetuning and can directly be utilized for inference with SOTA LLMs.
\paragraph{\textbf{Contributions}}%
1) We propose \MEdGellan, a novel hybrid decision-making framework that supports physicians in medical diagnosis by leveraging LLM-generated guidance. 2) We evaluate the framework using a complex, real-world clinical dataset. 3) We demonstrate that providing intermediate guidance on raw clinical inputs improves diagnostic performance.

\section{Related work}
\label{sec:related-work}
\paragraph{\textbf{Clinical decision-making}}  Recent years have seen a surge in the use of LLMs for clinical decision-making. Utilizing LLMs as independent decision-maker with only patient's data as input have been prolifically explored \citep{kim2024mdagents, wang2024drg, li2024exploring, zhu2024prompting, gao2024guiding}. Application of LLMs under interactive diagnostics system has also been explored \citep{wang2023chatcad, yunxiang2023chatdoctor}. \citet{li2023meddm} goes beyond using LLMs as just decision-makers and argued that LLMs, unlike doctors, lacks in differential medical knowledge and therefore provide suboptimal help to medical decision-making. As a solution, LLM-executable clinical guidance trees (CGT) were extracted from several diagnostic flowcharts. The CGTs bolster reasoning-capacity of the LLM when engaged in a multi-turn dialogue system. These are all fully automated systems and as such they can trigger automation bias and are incompatible with the high-stakes nature of the task.
\paragraph{\textbf{Prediction of discharge diagnosis}} The automatic prediction of discharge diagnoses from EHR data is an active area of research \citep{baksi-etal-2025-medcoder,boyle2023automated,edin2023automated, wu2025contrastive, barreiros2025explainable}.
While the prior works focus heavily on automated prediction of discharge diagnosis and raise the risk of \textit{separation of responsibilities} , \citet{sayin2025medsynenhancingdiagnosticshumanai} proposed an interactive chatbot tailored to assist clinicians in identifying correct diagnosis. In their setup,while the LLMs accesses entire clinical note, the physician has little access to the patient's data except the chief complaint of the patient and engages in a multi-turn dialogue with the LLM to collect relevant information. Thus, the authors avoid the potential risk of over-reliance on the machine. However, the LLM in their approach appears to be more like a customized oracle that answers patient-specific queries than a guidance-generator. 
\paragraph{\textbf{Framework to generate guidance}} 
% \ST{Mention that SLOG tackles SoR, otherwise what did we mention it for?}
\citet{banerjee2024learning} introduced a model called {\sc SLOG} to generate guidance from radiology reports. While their approach addresses the problem of {\em separation of responsibilities}, it relies on fine-tuning the underlying model and requires additional annotated data. In contrast, \MEdGellan operates without any need for fine-tuning or annotation, offering a more lightweight and scalable alternative.

\section{MedGellan}
\label{sec:medgel}

A patient's hospital stay is typically divided into several stages—such as the Emergency Department (ED), Intensive Care Unit (ICU), General Ward, and eventually, discharge. Diagnoses are usually recorded at the time of discharge. In this work, we use \MEdGellan to predict discharge diagnoses based on information available earlier in the clinical timeline, specifically the triage note and chest radiology report. The triage note, recorded upon admission to the ED, consists of initial clinical observations and vital signs. If necessary, a chest radiology test is conducted during the patient's ED stay. To reduce complexity and ensure a focused analysis, we restrict our study to patients who visited only the ED and underwent a single chest radiology test.

\MEdGellan is designed to assist physicians in predicting accurate discharge diagnoses. An overview of the proposed framework is shown in \cref{fig:pipeline}-right. In brief, \MEdGellan consists of two modules. The first module is powered by the \assistant—a state-of-the-art LLM—which takes the triage note and radiology report as input to generate clinical guidance. We ask the \assistant to \textit{observe} the triage information at first. Next, the \assistant is asked to look at the radiological report and update its assessment. Finally, based on its updated information, it generates a comprehensive guidance on the patient's health condition that would help the \physician to determine the ideal diagnosis. The second module is dedicated to the \physician. The \physician uses the guidance obtained from the first module and makes diagnosis for the patient. During this phase, the \physician is allowed to examine only the guidance, unlike the first module where the \assistant gets access to the raw information, i.e., the triage and the radiology report.

\section{Experimental Work}
\label{sec:expdet}

\paragraph{\textbf{Dataset}} We combine {\sc MIMIC-CXR} \citep{johnson2019mimic}, {\sc MIMIC-IV-ED} \citep{johnson2020mimic} and {\sc MIMIC-IV} \citep{johnson2024mimic} datasets for the chest radiographs, triage information, and diagnosis, respectively. \cref{tab:datamimic} summarizes the key components of each dataset.
{\sc Mimic-IV} is a comprehensive dataset containing health records of all patients who have visited the hospital. {\sc Mimic-IV-ED} is a subset focused specifically on emergency department data, while {\sc MIMIC-CXR} provides an extensive collection of chest radiograph images.
\begin{table}[!t]
    \centering
    \scriptsize
    \begin{tabular}{lccc}
        \toprule
        {\sc Dataset} & {\sc \# Pat} & {\sc \# Img} & {\sc \# Stays} \\
        \midrule
        \sc Mimic-CXR & $\sim$64k & $\sim$377k & NA \\
        \sc Mimic-IV-ED & NA & NA & $\sim$400k \\
        \sc Mimic-IV & $\sim$300k & NA & NA \\
        \bottomrule
    \end{tabular}
    \caption{Dataset statistics.}
    \label{tab:datamimic}
\end{table}%
We define a patient's diagnosis as the set of ICD codes assigned at the time of hospital discharge. Although diagnoses are also recorded when a patient leaves the ED, we do not predict them, as the radiology report's chart time often occurs after ED discharge—rendering such predictions unreliable. Instead, we focus on the final discharge diagnoses, represented with {\sc ICD-10} codes \citep{fung2020new}. After applying this filtering, we retained 1,366 unique hospital admissions, with each patient potentially assigned multiple {\sc ICD-10} codes.\footnote{In {\sc Mimic-IV}, these codes are listed in order of importance. Our framework, however, does not take this sequential ordering into account during prediction.}

\paragraph{\textbf{Models}}  As outlined in \Cref{sec:medgel}, the first task involves generating clinical guidance using an LLM, which is then reviewed by a physician. While the framework is designed to include a human physician, for preliminary experiments aimed at assessing its potential, we simulate the physician's role with an LLM.
We use Llama 3–70B \citep{grattafiori2024llama} as our \assistant model. For the \physicianLM model, we experiment with Llama 3 (8B and 70B) \citep{grattafiori2024llama}, Gemma 2–27B \citep{team2024gemma}, and Qwen2–72B \citep{team2024qwen2}.
All experiments are conducted using the Ollama framework.\footnote{\href{https://github.com/ollama}{https://github.com/ollama}}.

\paragraph{\textbf{Prompt}}  We provide the \assistant with both the triage note and the radiology report as input, prompting it to generate clinical guidance. Since the triage note is recorded prior to the radiological findings, we aim to preserve this temporal ordering in the prompting strategy. To do so, we adopt a Bayesian-inspired approach: the triage note is treated as prior knowledge, which the \assistant processes first. The radiology report is then introduced as new evidence, allowing the model to update its reasoning before generating the final guidance.
This strategy ensures temporal coherence between clinical events and avoids redundant or inconsistent reasoning. An example prompt used to generate guidance with the \assistant is shown below.

\begin{tcolorbox}[fonttitle=\small,fontupper=\scriptsize,colback=yellow!10, colframe=cyan!80!black, coltitle=black, title=Prompt for generating \textit{Guidance}]
    \label{bayes}
        {\tt You are an expert physician assistant trained to analyze patient records and generate a structured, evidence-weighted summary to aid in diagnosis. Your role is to synthesize information probabilistically, \textcolor{blue}{emphasizing prior observations (triage data) and new evidence (radiology findings) to refine the clinical understanding of the case.}\\
        \underline{\#\#\# In}p\underline{ut Data}:\\
        Patient records are provided in JSON format\:
        \textless Describe the columns \textgreater \\
        \underline{\#\#\# Ba}y\underline{esian-Ins}p\underline{ired Inference}:\\
        1. \textcolor{blue}{Prior Hypothesis} (Triage Data)  \\
           - Establish an initial clinical suspicion based on physiological indicators (vital signs, symptoms, and patient complaints).  \\
        2. \textcolor{blue}{Likelihood Adjustment} (Radiology Findings)  \\
           - Update the prior suspicion by assessing the radiology report.  \\
           - Weigh each new piece of evidence proportionally to its diagnostic importance.  \\
           - If imaging contradicts or reinforces the initial suspicion, adjust confidence accordingly. \\ 
        3. \textcolor{blue}{Posterior Summary} (Guidance for Diagnosis)  \\
           - Integrate both sources (triage + radiology) into a coherent, uncertainty-aware summary.\\  
           - Highlight most probable clinical concerns with confidence levels (e.g., "high likelihood of X, moderate possibility of Y").  \\
           - If findings are inconclusive, indicate potential differential diagnoses without committing to a single one.  \\
        \underline{\#\#\# Instructions}:\\
        - Use a Bayesian-inspired approach when synthesizing information:  \\
          - Begin with an initial assumption based on triage data.  \\
          - Adjust this assumption in light of radiology findings, emphasizing how new evidence modifies prior expectations. \\ 
          - Conclude with a refined summary, ensuring a logical progression of reasoning.  \\
        - Provide a structured, evidence-weighted summary of clinical observations. \\ 
        - Identify key abnormalities, trends, or risk factors while maintaining diagnostic neutrality.  \\
        - Use qualitative confidence levels (e.g., high, moderate, low) to reflect uncertainty in the summary.  \\
        - DO NOT provide final diagnoses or ICD-10 codes—\textbf{\textcolor{blue}{your role is to guide, not classify.
        }}}
\end{tcolorbox}

Once the guidance is generated, the \physicianLM—serving as a simulation of a real physician—takes this guidance as input and produces the final diagnosis. To evaluate the effectiveness of \MEdGellan, we compare its performance against two competitive baselines. In the first, we remove the \assistant from the pipeline and ask the \physicianLM to predict the final diagnosis using only the triage note. In the second, we again use the \physicianLM to predict the diagnosis, but this time provide both the triage note and the radiology report as input, without any intermediate guidance.

\paragraph{\textbf{Prediction}} We task the \physicianLM with predicting the corresponding ICD-10 codes for each diagnosis. These codes are structured hierarchically into three levels: chapter, category, and full code. Predicting the full codes is particularly challenging, as physicians may disagree on the finer-grained distinctions they entail \citep{sayin2025medsynenhancingdiagnosticshumanai}. To mitigate this ambiguity, we restrict the prediction task to the chapter and category levels.

\subsection{Results}
\label{sec:eval}

In~\Cref{tab:eval_1}, we compare the performance of the \physicianLM across three input settings: (i) using only the triage note, (ii) using both the triage note and the radiology report, and (iii) using the full \MEdGellan framework, which includes intermediate guidance. Predicting multiple ICD-10 codes for a single patient constitutes a challenging multi-label classification task. We evaluate performance using precision, recall, and $F_1$ score.
Across all \physicianLM variants used to simulate a real physician, the guidance provided by the \MEdGellan framework consistently outperforms the other two baselines in terms of recall and $F_1$. While we observe a slight drop in precision when guidance is included, the gains in recall and $F_1$—metrics more critical in high-stakes medical decision-making—indicate a favorable trade-off. This suggests that incorporating proper guidance enables the model to be more comprehensive in its predictions, thereby reducing the risk of false negatives.

\begin{table*}[!t]
    \centering
    \caption{Precision (Pr), Recall (Rec), and F1 scores across models and input types for Category and Chapter levels.}
    \label{tab:eval_1}
    \scalebox{0.7}{
    \begin{tabular}{lccccccccccccc}
            \toprule
            \sc Model & \sc Input &  \multicolumn{6}{c}{\sc Category} & \multicolumn{6}{c}{\sc Chapter} \\
             \cmidrule(lr){3-8} \cmidrule(lr){9-14}
            & & \multicolumn{3}{c}{\sc Micro} & \multicolumn{3}{c}{\sc Macro} & \multicolumn{3}{c}{\sc Micro} & \multicolumn{3}{c}{\sc Macro} \\
            \cmidrule(lr){3-5} \cmidrule(lr){6-8} \cmidrule(lr){9-11} \cmidrule(lr){12-14}
            & &  \sc Pr & \sc Rec & \sc F1 & \sc Pr & \sc Rec & \sc F1 & \sc Pr & \sc Rec & \sc F1 & \sc Pr & \sc Rec & \sc F1 \\
            \midrule
            \multirow{3}{*}{\sc LLAMA 8B} 
            & \sc Triage       & 0.11 & 0.03 & 0.05 & 0.12 & 0.04 & 0.05 & 0.64 & 0.24 & 0.35 & 0.65 & 0.27 & 0.36 \\
            & \sc Triage+Rad   & 0.07 & 0.01 & 0.02 & 0.07 & 0.02 & 0.03 & 0.57 & 0.15 & 0.24 & 0.58 & 0.17 & 0.25 \\
            & \sc Gui          & 0.19 & 0.09 & \textbf{0.12} & 0.19 & 0.11 & \textbf{0.13} & 0.64 & 0.38 & \textbf{0.48} & 0.66 & 0.42 & \textbf{0.48} \\
            \midrule
            \multirow{3}{*}{\sc LLAMA 70B} 
            & \sc Triage       & 0.0.43 & 0.12 & 0.19 & 0.48 & 0.16 & 0.22 & 0.74 & 0.28 & 0.41 & 0.78 & 0.32 & 0.43 \\
            & \sc Triage+Rad   & 0.40 & 0.12 & 0.18 & 0.46 & 0.15 & 0.21 & 0.72 & 0.27 & 0.39 & 0.77 & 0.31 & 0.41 \\
            & \sc Gui          & 0.33 & 0.17 & \textbf{0.22} & 0.35 & 0.21 & \textbf{0.24} & 0.65 & 0.40 & \textbf{0.50} & 0.67 & 0.44 & \textbf{0.50} \\
            \midrule
            \multirow{3}{*}{\sc Gemma2: 27B} 
            & \sc Triage       & 0.52 & 0.10 & 0.17 & 0.56 & 0.13 & 0.20 & 0.81 & 0.23 & 0.36 & 0.83 & 0.27 & 0.38 \\ 
            & \sc Triage+Rad   & 0.53 & 0.08 & 0.14 & 0.55 & 0.11 & 0.18 & 0.78 & 0.19 & 0.31 & 0.79 & 0.22 & 0.34 \\
            & \sc Gui          & 0.42 & 0.12 & \textbf{0.19} & 0.43 & 0.16 & \textbf{0.22} & 0.72 & 0.30 & \textbf{0.42} & 0.73 & 0.35 & \textbf{0.44} \\
            \midrule
            \multirow{3}{*}{\sc Qwen2: 72B} 
            & \sc Triage       & 0.20 & 0.05 & 0.08 & 0.19 & 0.06 & 0.09 & 0.67 & 0.25 & 0.36 & 0.69 & 0.27 & 0.37 \\
            & \sc Triage+Rad   & 0.23 & 0.07 & 0.11 & 0.25 & 0.09 & 0.13 & 0.60 & 0.26 & 0.36 & 0.63 & 0.29 & 0.37 \\
            & \sc Gui          & 0.24 & 0.09 & \textbf{0.13} & 0.25 & 0.11 & \textbf{0.14} & 0.59 & 0.30 & \textbf{0.40} & 0.61 & 0.33 & \textbf{0.41} \\
            \bottomrule
    \end{tabular}}
\end{table*}

\section{Conclusion and Future Work}
\label{sec:conclusion}

In this study, we introduced \MEdGellan, an efficient framework for hybrid medical decision-making that requires no finetuning or annotations.
Our preliminary experiments demonstrate how ``guidance'' obtained with \MEdGellan can enhance the quality of predicted diagnoses over the sole usage of raw inputs, including triage information or radiology reports.
In future work, we plan to investigate the impact of guidance when presented to human physicians and to extend \MEdGellan to utilize rich non-textual information, such as the radiology images themselves, along with text data.

\bibliographystyle{plainnat}
\bibliography{paper}
\end{document}